\documentclass[10pt,twocolumn,letterpaper]{article}

\usepackage{iccv}              

\definecolor{iccvblue}{rgb}{0.21,0.49,0.74}
\usepackage[pagebackref,breaklinks,colorlinks,allcolors=iccvblue]{hyperref}

\usepackage{amsmath}
\usepackage{amssymb}
\usepackage{mathtools}
\usepackage{amsthm}
\usepackage{graphicx}

\usepackage{caption}
\usepackage{subcaption}

\usepackage{hyperref}
\usepackage{url}
\usepackage{booktabs} 
\usepackage{wrapfig} 
\newcommand{\tablestyle}[2]{\setlength{\tabcolsep}{#1}\renewcommand{\arraystretch}{#2}\centering\footnotesize}

\def\module{Isolated Identity Adapter }

\title{PersonalVideo: High ID-Fidelity Video Customization without \\ Dynamic and Semantic Degradation}

\author{%
Hengjia Li$^{1}$ \and Haonan Qiu$^{2}$ \and Shiwei Zhang$^{3,*}$\and Xiang Wang$^{3}$\and Yujie Wei$^{3}$\and Zekun Li$^{3}$\and Yingya Zhang$^{3}$\and Boxi Wu$^{1}$\and Deng Cai$^{1}$ 
\\\\
 $^1$Zhejiang University\qquad
 $^2$Nanyang Technological University\qquad
 $^3$Alibaba Group
 \\
 {\tt\small lhjzju@zju.edu.cn},\quad
 {\tt\small zhangjin.zsw@alibaba-inc.com}\\
 \vspace{-0.6em} \\ 
 Project page: \url{https://personalvideo.github.io/}
}

\usepackage[misc]{ifsym}
\newcommand\blfootnote[1]{
    \begingroup
    \renewcommand\thefootnote{}\footnote{#1}
    \addtocounter{footnote}{-1}
    \endgroup
}
\usepackage[hang]{footmisc}

\begin{document}

\twocolumn[{
\renewcommand\twocolumn[1][]{#1}
\maketitle
\begin{center}
    \centering
    \includegraphics[width=.95\textwidth]{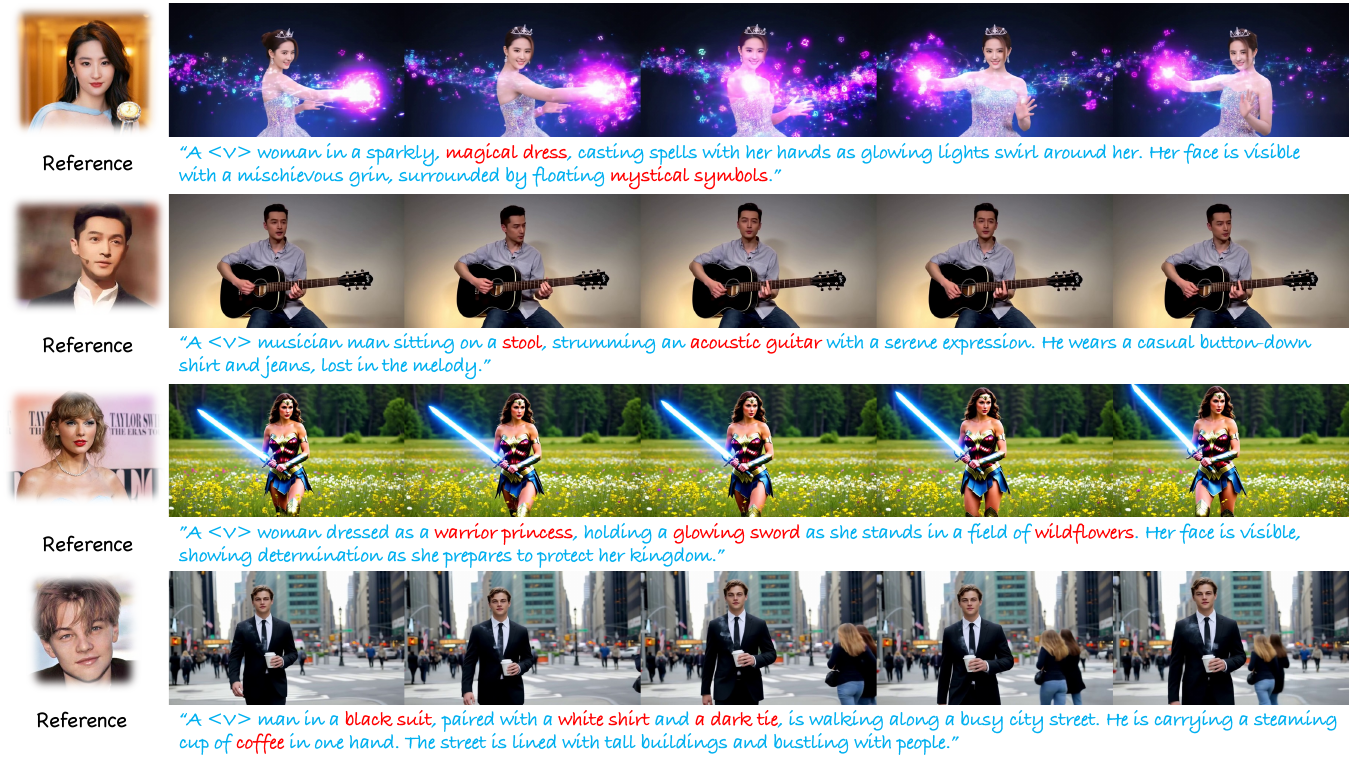}
    \vspace*{-.1cm}
        \captionof{figure}{
        \textbf{Results of PersonalVideo.} Given the reference images of a specific identity, PersonalVideo can generate high ID-fidelity videos with promising motion dynamics and prompt following.
      }
\label{fig:teaser}
\end{center}
\vspace*{.1cm}
}]

{
    \blfootnote{
        * Project Leader        }
}

\begin{figure*}[t]
\centering
\includegraphics[width=.9\linewidth]{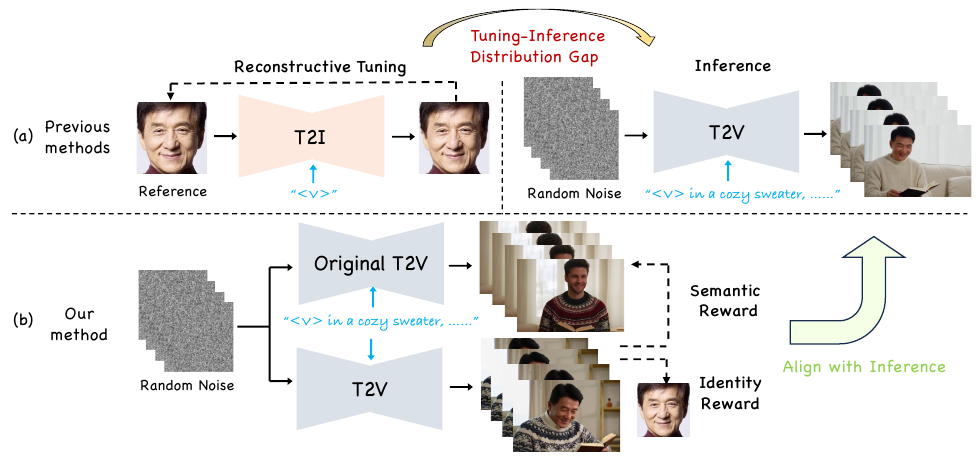}
\vspace{-0.4cm}
\caption{
\textbf{Analysis of the tuning-inference gap.} Previous T2V customization supervises the tuning process via reconstructing images on T2I models, suffering from a tuning-inference gap. Differently, we aim to directly apply the supervision on generated videos, which aligns with inference and bridges the gap.}
\label{fig:image}
\vspace{-0.4cm}
\end{figure*}
\begin{abstract}
The current text-to-video (T2V) generation has made significant progress in synthesizing realistic general videos, but it is still under-explored in identity-specific human video generation with customized ID images. The key challenge lies in maintaining high ID fidelity consistently while preserving the original motion dynamic and semantic following after the identity injection. 
Current video identity customization methods mainly rely on reconstructing given identity images on text-to-image models, which have a divergent distribution with the T2V model. This process introduces a tuning-inference gap, leading to dynamic and semantic degradation. 
To tackle this problem, we propose a novel framework, dubbed \textbf{PersonalVideo}, that applies a mixture of reward supervision on synthesized videos instead of the simple reconstruction objective on images. Specifically, we first incorporate identity consistency reward to effectively inject the reference's identity without the tuning-inference gap. Then we propose a novel semantic consistency reward to align the semantic distribution of the generated videos with the original T2V model, which preserves its dynamic and semantic following capability during the identity injection. 
With the non-reconstructive reward training, we further employ simulated prompt augmentation to reduce overfitting by supervising generated results in more semantic scenarios, gaining good robustness even with only a single reference image.
Extensive experiments demonstrate our method's superiority in delivering high identity faithfulness while preserving the inherent video generation qualities of the original T2V model, outshining prior methods. 
\end{abstract}
\section{Introduction}

Recently, there has been considerable scholarly interest in text-to-video (T2V) generation~\citep{guo2023animatediff,wang2023modelscope,chen2024videocrafter2,videoworldsimulators2024}, which allows for the production of videos from user-defined textual descriptions. However, the identity-specific customization of high-fidelity human videos has not been fully explored. It aims to customize a wide variety of engaging videos using a few users' photos, allowing for personalized content creation that features them in different actions, scenes, or styles while maintaining high ID fidelity. Without the need for complex scene construction and tedious post-production special effects, this convenient way of video creation also has great potential in the film and television industry.

Identity customization has achieved significant advancements in the field of text-to-image (T2I)~\citep{gal2022image, ruiz2023dreambooth, ye2023ip, li2024photomaker, wang2024instantid, guo2024pulid}. Typically, these methods use a reconstructive approach on provided reference images to inject the identity into the pre-trained T2I model during the customization. However, directly employing this strategy in video customization~\citep{wei2023dreamvideo, ma2024magic} will lead to unsatisfied results due to two notable challenges: 

(1) \textbf{Preserving inherent motion dynamics and semantic following.}
Existing video customization methods~\citep{ma2024magic, wei2023dreamvideo} naively use image reconstruction supervision during tuning to model a customized T2I prior, which is then injected into the T2V model to generate identity-specific videos during inference. However, the distribution of the pre-trained T2V model often deviates from that of the pre-trained T2I model, which brings a tuning-inference gap as shown in \cref{fig:image}.  Since tuning on limited static images will significantly shift the video prior of the pre-trained T2V model, it leads to a dynamic and semantic degradation, making the generated videos tend to appear static and fail to follow the given prompts. 

(2) \textbf{Inserting consistent identity with high fidelity.} For reconstruction-based video customization, the tuning-inference gap also brings challenges for ID-fidelity. As humans are sensitive to facial features, higher fidelity, and consistent identity are required in customized videos. To achieve identity consistency while preserving the model's dynamics and semantics, traditional reconstructive video customization methods~\citep{wei2023dreamvideo, he2023animate, chefer2024still} often require more images, even additional video input, to inject identity without overfitting, which brings great inconvenient for users.

To address these challenges, we propose a novel framework, dubbed \textbf{PersonalVideo}, for ID-specific video generation that can achieve high ID fidelity and maintain original motion dynamics and semantic following with only a few images of an identity given. 
Different from previous methods with a reconstructing objective on T2I models, our core insight is \textit{applying the direct reward to videos generated by the T2V model} thus bridging the tuning-inference gap as shown in \cref{fig:image}. Specifically, we integrate feedback from a mixture of differentiable reward models to inject an identity consistent with the reference while preserving the original model's dynamic and semantic following capability.

Inspired by the successful experience of the identity supervision~\citep{wang2021towards, richardson2021encoding} in encoder-based T2I customization~\citep{gal2024lcm,guo2024pulid}, we propose to apply Identity Consistency Reward (ICR) to a randomly selected frame in the generated video, leveraging the similarity feedback from an identity recognition model to consistently inject the reference’s identity without the tuning-inference gap. However, applying only ID reward still suffers from degraded dynamic and semantic following capabilities, which is caused by the inherent distribution shift introduced by the limited number of static images during ID injection. To address this shift, we further propose a Semantic Consistency Reward (SCR) to preserve the original model's semantic distribution, thereby mitigating the degradation of dynamic and semantic following capabilities. Leveraging the semantic reward model, we can evaluate the image-text correspondence of the sampled video frames and then align the score distribution between the original video and the target video, thus effectively preserving the dynamics and semantics of the original generator without affecting the injection of the identity.


With the non-reconstructive reward supervision, we also employ simulated prompt augmentation, which is independent of provided reference images and supervises generated results in more semantic scenarios. Unlike the reconstructive training method, they do not correspond to each reference nor are they limited by the number of reference images, effectively mitigating overfitting and demonstrating strong robustness, even when only a single reference image is available. 


To achieve the identity injection with preserved motion dynamics and semantic following, it is also essential to design the learnable modules. Based on the observations for the denoising steps, we further propose an \module injected in the only later denoising step, which minimizes the impact on the original video’s dynamic. Qualitative and quantitative experiments demonstrate that our \textbf{PersonalVideo} achieves high ID fidelity and effectively preserves the original T2V model's capabilities. Benefiting from the scalability of T2V models, it also supports any style-specific fine-tuned model, which offer valuable flexibility for abundant creation in the AIGC community. Our contributions are summarized as follows:
\begin{itemize}
\item We introduce a novel framework, dubbed \textbf{PersonalVideo}, for video personalization with static images. To bridge the tuning-inference gap, we propose to apply the direct reward feedback to generated videos, achieving high ID-fidelity without dynamic and semantic degradation.

\item We propose a novel semantic consistency reward to effectively preserve the original model's semantic distribution without affecting the injection of the identity.

\item With the non-reconstructive supervision, we introduce simulated prompt augmentation which is independent of the reference images and robustly improves ID-fidelity, even for just a single image.

\end{itemize}
\section{Related Work}
\textbf{Text-to-Video Generation.} The topic of T2V generation has attracted considerable interest among researchers for a long time.
Recently, the utilization of diffusion models has become predominant in the realm of T2V tasks. VDM~\citep{ho2022video} stands as the pioneer that first leverages a diffusion model for T2V generation. 
Subsequently, Make-A-Video~\citep{singer2022make} and Imagen Video~\citep{ho2022imagenvideo} were proposed to generate high-resolution videos in pixel space.
To save computational resources, various frameworks have been developed to perform a latent denoising process~\citep{zhou2022magicvideo,he2022latent,wang2023modelscope,blattmann2023align,wang2023lavie}. Although these methods can produce high-quality generic videos by pre-training on large-scale text-video pair datasets, it remains challenging to synthesize id-specific videos.

\noindent
\textbf{Text-to-Image Customization.} In the field of T2I, a lot of approaches~\citep{gal2022image, li2024photomaker, gal2024lcm, valevski2023face0, xiao2024fastcomposer, ma2024subject, peng2024portraitbooth, li2023few, li2024unihda} have emerged for ID customization. As a seminal work, Textual Inversion~\citep{gal2022image} represents the user-provided identity with a specific token embedding of a frozen T2I model. For better ID fidelity, DreamBooth~\citep{ruiz2023dreambooth} optimizes the original model, where efficient fine-tuning techniques such as LoRA~\citep{hu2021lora} can also be applied. On the other hand, encoder-based methods aims to directly inject ID into the generation process.
IP-Adapter~\citep{ye2023ip} and InstantID~\citep{wang2024instantid} focus on adapting encoders that extract ID-relevant information. 
PuLID~\citep{guo2024pulid} suggests optimizing an ID loss between the generated and reference images in a more accurate setting.

\begin{figure*}[t]
\centering
\includegraphics[width=.9\linewidth]{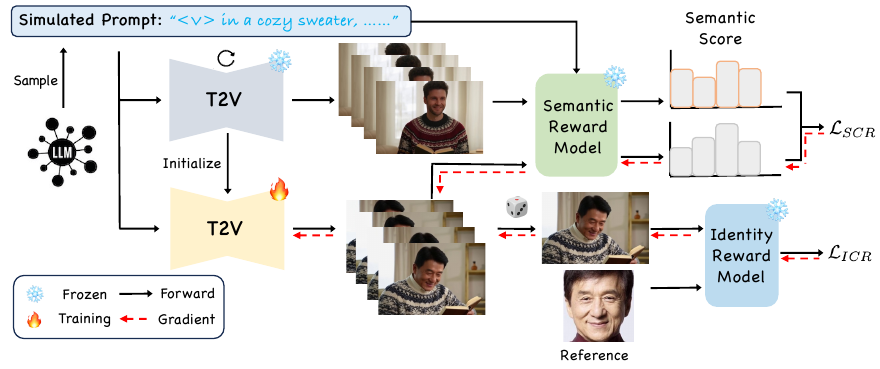}
\vspace{-0.3cm}
\caption{\textbf{Overview of the framework of PersonalVideo.} To bridge the tuning-inference gap, we directly apply reward supervision on generated videos starting from pure noises, including identity consistency reward with the reference and semantic consistency reward with the original video. During the optimization, we adopt simulated prompt sampled from the Large Language Model to supervise generated results in more semantic scenarios.
}
\label{fig:method}
\vspace{-0.5cm}
\end{figure*}

\noindent
\textbf{Text-to-Video Customization.} The T2V customization presents further challenges compared to the T2I customization due to the temporal motion dynamics involved in videos. Currently, only limited works~\citep{ma2024magic, he2024id, wu2024customcrafter, wei2024dreamvideo, yuan2024identity} have undertaken early investigations into this area. MagicMe~\citep{ma2024magic} adopts an ID module based on extended Textual Inversion. However, training under T2I reconstruction supervisiondeviates from the T2V inference, leading to inferior ID fidelity and model degradation. 
DreamVideo~\citep{wei2023dreamvideo} customizes the identity given a few images, but the inconvenience lies in the fact that it necessitates additional videos to provide motion patterns. ID-Animator~\citep{he2024id} proposes to encode ID-relevant information with a face adapter, which requires thousands of high-quality human videos for fine-tuning, thereby incurring significant costs associated with dataset construction and model training.

\section{Preliminaries}
\vspace{-0.1cm}
Text-to-video diffusion models (T2V)~\citep{blattmann2023align, guo2023animatediff, wang2023modelscope, wang2024animatelcm, kong2024hunyuanvideo} are tailored for generating videos by adapting image diffusion models to handle video data. 
Specifically, the diffusion model $\epsilon_{\theta}$ aims to predict the added noise $\epsilon$ at each timestep $t$ based on text condition $c$, where $t \in \mathcal{U}(0, 1)$ is normalized. The training objective can be simplified as a noise-prediction loss:
\begin{align}
    \mathcal{L}_{\text{diff}} = \mathbb{E}_{z, c, \epsilon \sim \mathcal{N}(0, \mathbf{\mathrm{I}}), t}\left[\left\| \epsilon - \epsilon_\theta\left(z_t, \tau_\theta(c), t\right)\right\|_2^2\right],
\label{eq:diffusion_loss}
\end{align}
where $z \in \mathbb{R}^{B \times F \times H \times W \times C}$ is the latent code of video data with $B, F, H, W, C$ being batch size, frame, height, width, and channel, respectively. $\tau_\theta$ presents a pre-trained text encoder. A noise-corrupted latent code $z_t$ from the ground-truth $z_0$ is formulated as $z_t = \alpha_t z_0 + \sigma_t \epsilon$,
where 
$\alpha_t$ and $\sigma_t$ are hyperparameters to control the diffusion process. 


\section{Methodology}

\subsection{Non-Reconstructive Reward Framework}
\label{sec:recon}
Current T2V customization typically adopts a reconstruction approach to train a customized T2I prior on provided images and inject it into the pre-trained T2V models to generate identity-specific videos. However, it leads to a tuning-inference gap due to the misaligned distribution between reference images during tuning and generated videos in inference time, which brings inferior ID fidelity and degradation of inherent motion dynamics and semantic following.  

To bridge the gap, as shown in \cref{fig:method}, we propose a non-reconstructive framework to directly apply reward on the generated videos for high ID fidelity without dynamic and semantic degradation. Inspired by the successful experience of the identity supervision~\citep{wang2021towards, richardson2021encoding} in encoder-based T2I customization~\citep{gal2024lcm,guo2024pulid}, we propose to apply Identity Consistency Reward (ICR) to the generated video. During the tuning time, we start from pure noise instead of noised references in the previous reconstructive methods. Then we directly supervise the a randomly selected frame from the generated video to mimic the identity in the generated videos with that in references using ID similarity reward, which closely aligns with human perception and the distribution of the real world. 

Specifically, we use pre-trained ID recognition model~\citep{deng2019arcface} $\mathcal{R}_{id}$ to precisely extract the ID embeddings of the references and the random $i$-th frame of the generated video. Then we minimize the cosine similarity of them to align the identity effectively and directly. Here we crop the faces and adopt image augmentation techniques like color jitter to make it more robust for limited references. 
Formally, our training objective is:
\begin{equation}
    \label{eq:objective_id}
    \mathcal{L}_{\text{ICR}} = \mathbb{E}_{i, c\sim p(c)}\left[\texttt{CosSim}\left(\mathcal{R}_{id}(I_{ref}), \mathcal{R}_{id}(G_{\mathcal{T}}(z_T, c, i))\right)\right],
\end{equation}
where $I_{ref}$ are the reference images, $G_{\mathcal{T}}$ is the target trained diffusion model with the VAE decoder, and $c$ is the text prompt with the specific keyword.

\subsection{Semantic Consistency Reward}
While the non-reconstructive ICR addresses the tuning-inference gap in existing methods, the distribution gap introduced by the limited static references still remains, which makes it challenging to preserve the original dynamics and semantics. To address this issue, we further introduce a novel Semantic Consistency Reward (SCR) to maintain the original model's semantic distribution, effectively alleviating the decline in dynamic and semantic following. Specifically, we randomly sample a batch of $M$ frames from the video generated by original and target model. Then we leverage the semantic reward model $\mathcal{R}_{sem}$ to evaluate the image-text correspondence of the sampled video frames
\begin{equation}
\begin{gathered}
V_c^\mathcal{S} = \texttt{Softmax}(\{\mathcal{R}_{sem}(G_{\mathcal{S}}(z_T, c, i)\}_{i=1}^M)), \\
V_c^\mathcal{T} = \texttt{Softmax}(\{\mathcal{R}_{sem}(G_{\mathcal{T}}(z_T, c, i)\}_{i=1}^M)),
\end{gathered}
\end{equation}
where $G_{\mathcal{S}}$ is the frozen original diffusion model with the VAE decoder.
Finally, we align the score distribution between the original video and the target video with KL divergence,
\begin{equation}
\begin{gathered}
\mathcal{L}_{\text{SCR}} = \mathbb{E}_{c\sim p(c)}D_{KL}(V_c^T||V_c^S).
\end{gathered}
\end{equation}
Through the distribution alignment with the original generator, we can effectively preserve its dynamics and semantics without affecting the injection of the identity. 

Thus, the full learning objective is defined as:
\begin{equation}
\begin{gathered}
\mathcal{L}_{\text{train}} = \mathcal{L}_{\text{ICR}} + \mathcal{L}_{\text{SCR}}.
\end{gathered}
\end{equation}

\begin{figure}
\centering
\includegraphics[width=.85\linewidth]{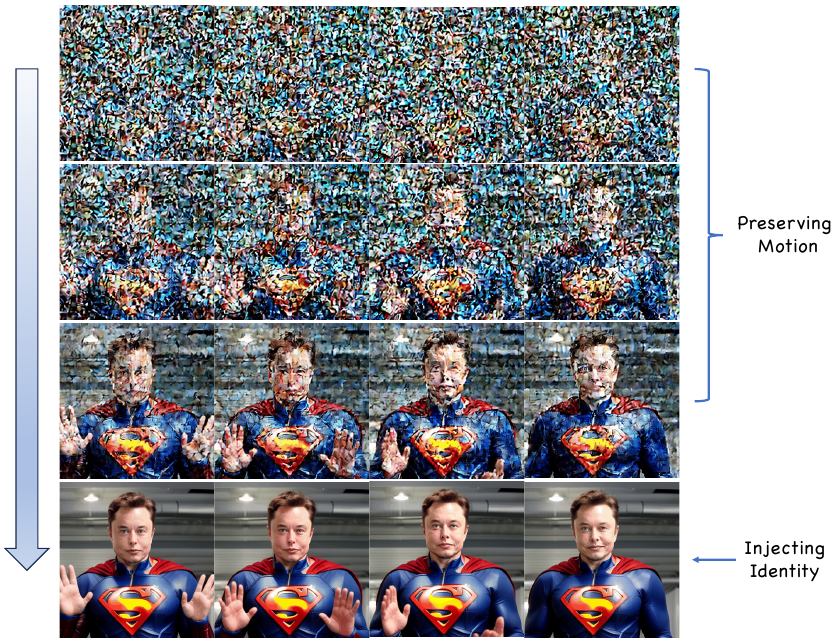}
\vspace{-0.3cm}
\caption{\textbf{Visualization of the video denoising steps.} The motion of the person, \eg, his hand, is formed in early stages of the denoising process. the later steps focus on the recovering of the detailed appearance.}
\label{fig:denoise}
\vspace{-0.4cm}
\end{figure}

\subsection{Simulated Prompt Augmentation}
\label{sec:aug}
To further enhance the robustness of the customization, we introduce simulated prompt augmentation. During the customization process, traditional reconstruction frameworks can only utilize prompts that describe the reference image, which limits the model's generalization capabilities. Benefiting from the non-reconstructive framework, we can incorporate numerous reference-irrelevant prompts during the optimization, \eg, \textit{`V' playing the violin} and \textit{`V' smiling on the beach}. Specifically, we leverage the Large Language Model to create 50 prompts as our simulated prompts with various motion, appearance, and backgrounds and randomly select the prompts during the customization. They align well with actual test scenarios to mitigate overfitting with strong robustness, even when only a single reference image is available. 

\subsection{\module}
To achieve the identity injection with preserved motion dynamics and semantic following, it is also essential to design the learnable modules. Based on the observations for the denoising steps shown in \cref{fig:denoise}, we find that the motion of the person is typically formed in the early denoising step. During these steps, the model tends to restore the layout~\citep{cao2023masactrl} and motion. In contrast, the later steps focus on recovering of the detailed appearance of the objects~\citep{patashnik2023localizing}. Therefore, we propose to isolatedly inject the identity only in the later denoising steps during training and inference time, to reduce the influence on the motion generation and mitigate the distribution shift caused by static reference images. Formally, it adopts the residual path of two low-rank matrices including a down block $A^{\text{down}}\in\mathbb{R}^{d\times r}$ and up block $A^{\text{up}}\in\mathbb{R}^{r\times k}$.
Formally, the updated parameter matrices are  
\begin{equation}
    \tilde{W} = W + \Delta W = W + A^{\text{down}} A^{\text{up}},
\label{eq:LoRA}
\end{equation}
for all $W$ in the layers of query, key, and value.

\begin{figure*}[t]
\centering
\includegraphics[width=.97\linewidth]{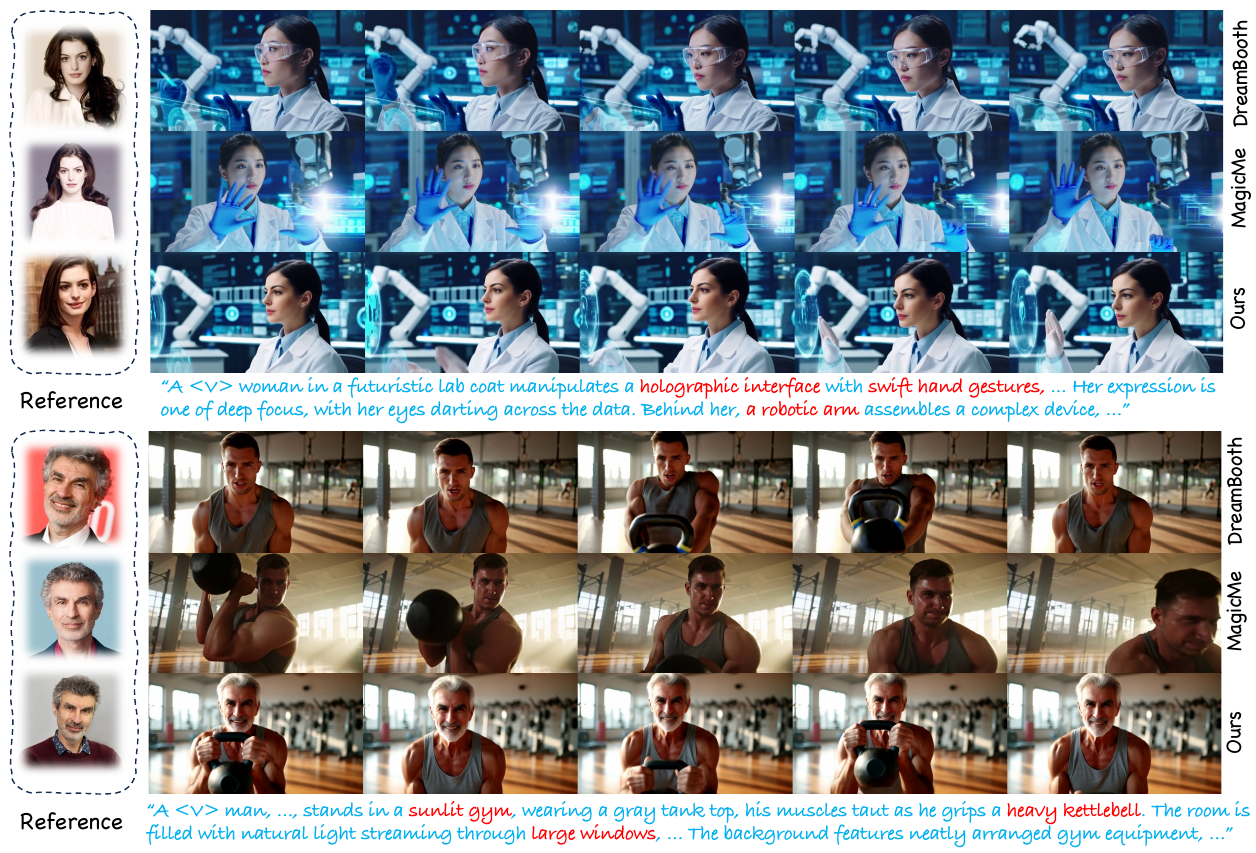}
\vspace{-0.2cm}
\caption{\textbf{Qualitative comparison for a few references.} As observed, both Dreambooth and MagicMe suffer from inferior ID fidelity. In contrast, our PersonalVideo maintains high ID fidelity and preserve the original motion dynamics and semantic following, significantly surpassing others.}
\label{fig:new1}
\vspace{-0.2cm}
\end{figure*}

\begin{table*}[ht]
    \tablestyle{3pt}{1.15}
    \centering
    \begin{tabular}{ccccccc} 
        \toprule
\textbf{Method} & \textbf{Face Sim.} ($\uparrow$) & \textbf{Dyna. Deg.} ($\uparrow$) & \textbf{FVD} ($\downarrow$)& \textbf{T. Cons.} ($\uparrow$) & \textbf{CLIP-T} ($\uparrow$)& \textbf{CLIP-I} ($\uparrow$)\\
\midrule
DreamBooth &42.62 &13.86 &1325.89 &0.9919   &26.26 &44.27 \\
MagicMe &50.51 &11.88 &1336.73 &0.9928  &25.48 &73.03 \\
IDAnimator  &43.88 &14.33 &1538.44   &0.9912 &24.33 &50.23 \\
ConsisID &53.22 &15.22  &1622.21  &0.9923 &25.39 &74.58\\
\textbf{PersonalVideo} &\textbf{62.35} &\textbf{17.80} &\textbf{1272.32} &  \textbf{0.9935} &\textbf{26.30} & \textbf{76.48} \\
        \bottomrule
    \end{tabular}
\vspace{-0.2cm}
\caption{\textbf{Quantitative comparison.} The metrics cover the ability to achieve high ID fidelity (\ie, Face Similarity and CLIP-I), dynamic degree, text alignment (\ie, CLIP-T), distribution distance (\ie, FVD), and temporal consistency. 
}
\label{tab:result}
\vspace{-10pt}
\end{table*}

\begin{figure*}[t]
\centering
\includegraphics[width=\linewidth]{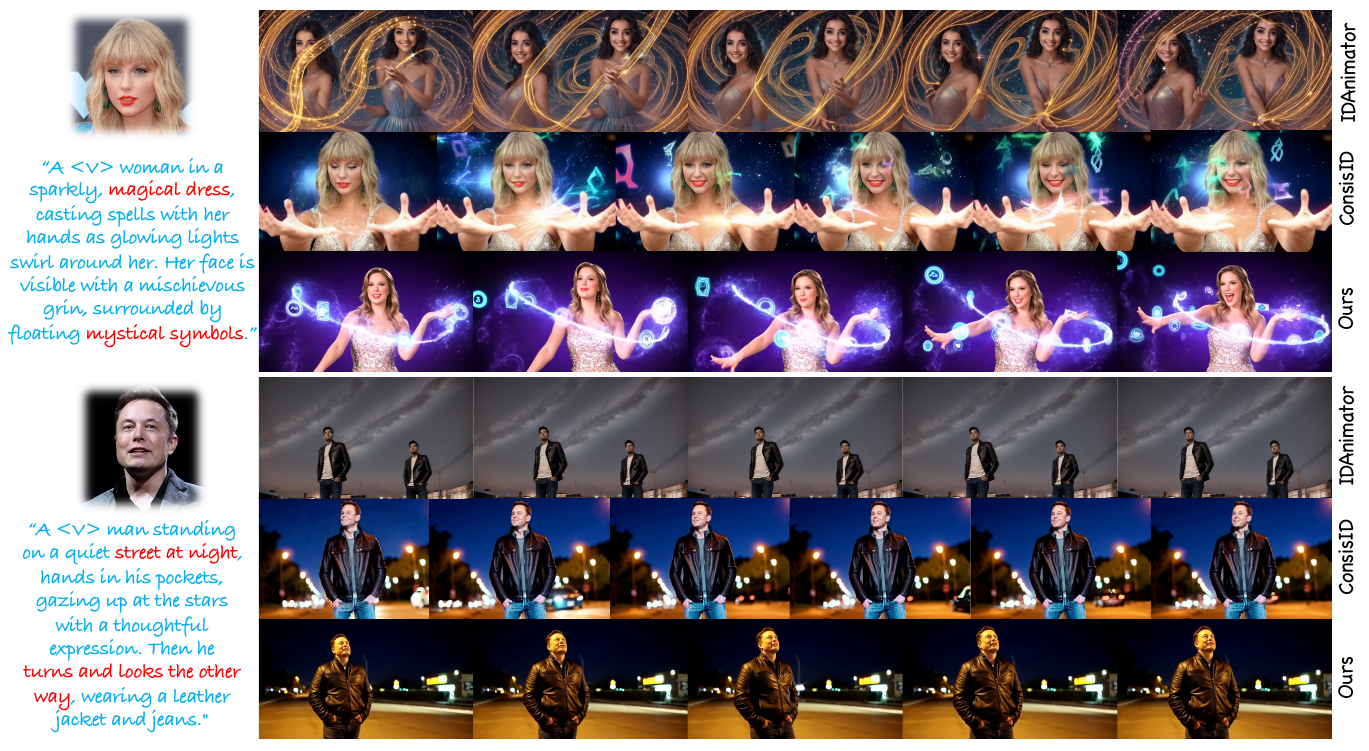}
\vspace{-0.6cm}
\caption{\textbf{Qualitative comparison for a single reference.} As observed, the baseline methods not only demonstrate suboptimal identity fidelity but also suffer from compromised dynamic motions and insufficient video quality. In comparison, our PersonalVideo maintains higher ID fidelity without dynamic and semantic degradation, which is consistent with \cref{fig:new1}.}
\label{fig:new2}
\vspace{-0.4cm}
\end{figure*} 
\vspace{-0.1cm}
\section{Experiments}

\begin{figure}[t]
\centering
\includegraphics[width=.8\linewidth]{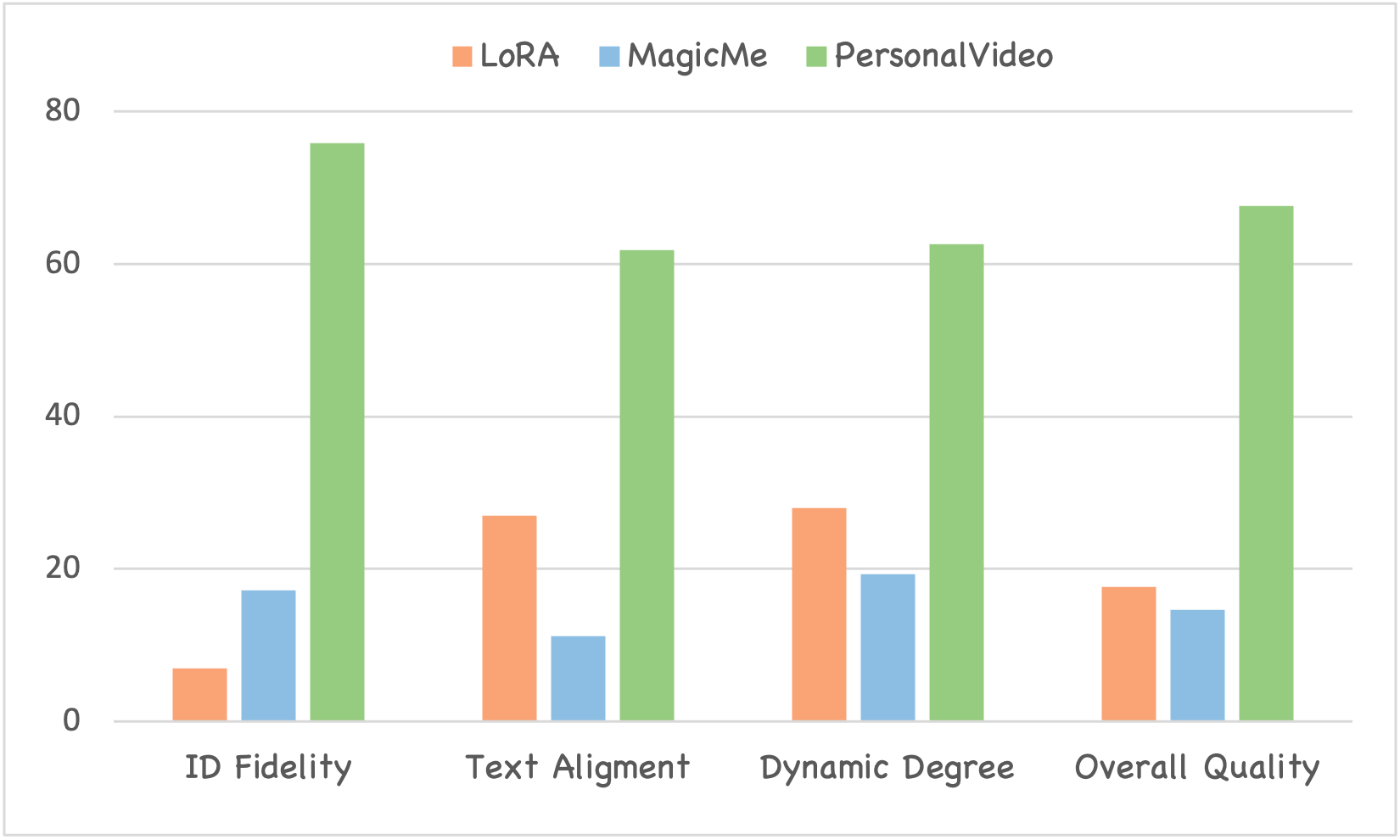}
\vspace{-0.1cm}
\caption{\textbf{User study.} Our PersonalVideo achieves the best human preference compared with DreamBooth and MagicMe.}
\label{fig:user}
\vspace{-0.3cm}
\end{figure}

\begin{figure*}[t]
\centering
\includegraphics[width=.85\linewidth]{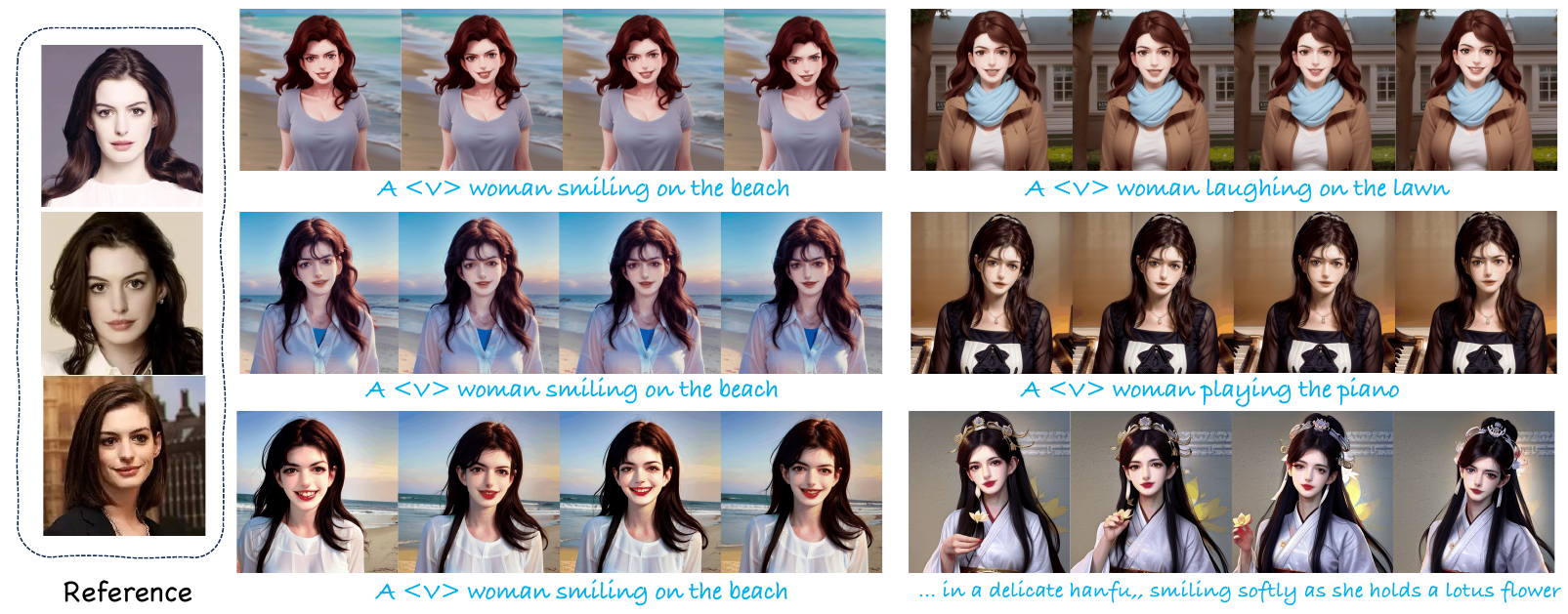}
\vspace{-0.3cm}
\caption{\textbf{Compatibility with customized style LoRAs,} \ie, RCNZ Cartoon 3D, GuoFengRealMix, and GuoFeng.}
\label{fig:style}
\vspace{-0.5cm}
\end{figure*}

\subsection{Experimental Settings}
To demonstrate the effectiveness of our method, we utilize a DiT-based model, HunyuanVideo~\cite{kong2024hunyuanvideo}, and a UNet-based model AnimateDiff as our T2V generation models. We use ResNet-100~\citep{he2016deep} backbone pre-trained on Glint360K~\citep{an2021partial} dataset as the identity reward model and HPSv2~\citep{wu2023human} as the semantic reward model. To demonstrate the superiority of PersonalVideo, we compare it with MagicMe~\citep{ma2024magic}, a recent identity-specific T2V customization method, as well as Dreambooth with LoRA~\citep{ruiz2023dreambooth, hu2021lora}. Additionally, we compare the results with encoder-based methods like IDAnimator~\citep{he2024id} and ConsisID~\cite{yuan2024identity}. We present the details of baselines and training process in Appendix.

\subsection{Qualitative Results}
We provide a qualitative comparison between PersonalVideo and baselines. As shown in \cref{fig:new1}, both Dreambooth and MagicMe suffer from inferior ID fidelity, due to the tuning-inference gap.  In contrast, our PersonalVideo achieves high ID fidelity and preserves the original motion dynamics and semantic following.  
To verify the robustness of our method, we also conduct the experiments for only a single image reference and compare with encoder-based methods. As shown in \cref{fig:new2}, they not only demonstrate suboptimal identity fidelity but also suffer from compromised dynamic motions and insufficient video quality. The results further underscore the superiority of PersonalVideo, with promising robustness to achieve high ID fidelity and preserve motion dynamics and semantic following. 
To demonstrate the generalization of our method, we also conduct the experiments on a UNet-based model, Animatediff, in the appendix.

\subsection{Quantitative Results}
We present the quantitative results in \cref{tab:result} and evaluate 1000 generated videos for 20 identities with 50 prompts from these perspectives: (1) Face Similarity: the ID cosine similarity to evaluate ID fidelity, with ID embeddings extracted using Antelopev2~\citep{deng2019arcface}, which is different from the face recognition models in our framework. (2) Dynamic Degree: we use VBench~\citep{huang2024vbench}, an effective benchmark to compute video dynamics. Besides, we use well-known metrics for video evaluation. As observed, DreamBooth suffers from inferior face similarity due to the tuning-inference gap. MagicMe gets better face similarity yet degraded dynamics and text alignment. 
In contrast, our PersonalVideo significantly surpasses previous methods, especially for face similarity and dynamic degree. It achieves high ID fidelity and effectively preserves the ability of the original T2V model, which is consistent with the qualitative results.

\subsection{User Study}
\label{sec:user}

To further assess the effectiveness of our approach, we perform a human evaluation comparing our method with existing T2V identity customization techniques. We invite 15 people to review 50 sets of generated video results. For each set, we provide reference images alongside videos created using the same seed and text prompt across different methods. We evaluate the quality of the generated videos on four criteria: Identity Fidelity (the resemblance of the generated object to the reference image), Text Alignment (how well the video corresponds to the text prompt),  Dynamic Degree (the dynamic degree of motion in the video), and Overall Quality (the overall satisfaction of users with the video quality). As illustrated in \cref{fig:user}, our PersonalVideo receives significantly higher user preference across all evaluation metrics, demonstrating its effectiveness.

\subsection{Compatibility with LoRAs}
We also showcase that our framework demonstrates excellent compatibility with existing fine-grained condition modules, such as style LoRAs. As shown in \cref{fig:style}, We use the Civitai community models to show that it operates effectively with these weights, even though it was not specifically trained on them. The first row displays the results from the RCNZ Cartoon 3D model~\citep{rcnzcartoon2023}, while the second row and the third row highlight the outcomes from the GuoFeng RealMix~\citep{guofeng2023} and GuoFeng model~\citep{realmix2023}. Our approach consistently delivers reliable facial preservation and effective motion generation, which has great compatibility with these customized style LoRAs.

\begin{figure}[t]
\centering
\includegraphics[width=.95\linewidth]{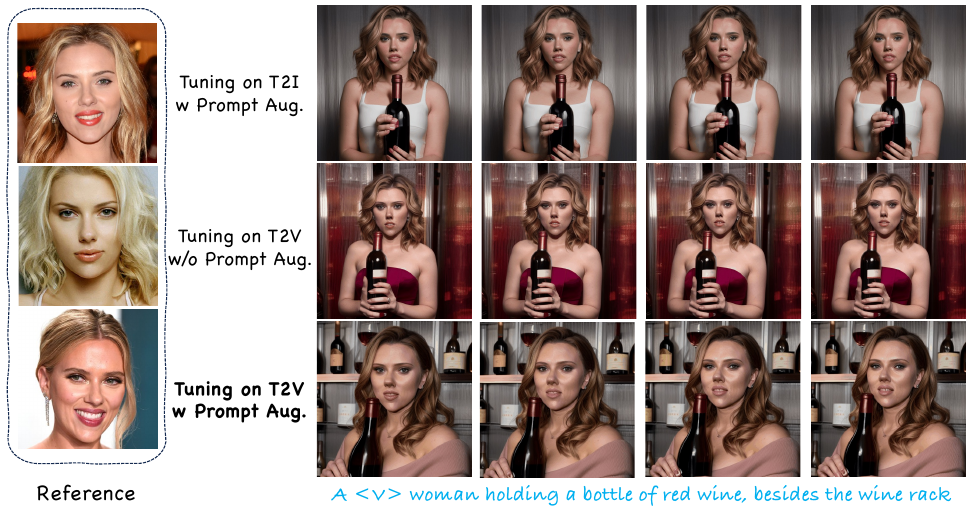}
\vspace{-0.3cm}
\caption{\textbf{Ablation study for the non-reconstructive training and simulated prompt augmentation.} As observed, tuning on the T2I model suffers from inferior ID fidelity and blurred background. Besides, tuning without prompt augmentation degrades the semantic following, \ie, \textit{the wine rack}.}
\label{fig:text}
\vspace{-0.3cm}
\end{figure}

\begin{figure}[t]
\centering
\includegraphics[width=.95\linewidth]{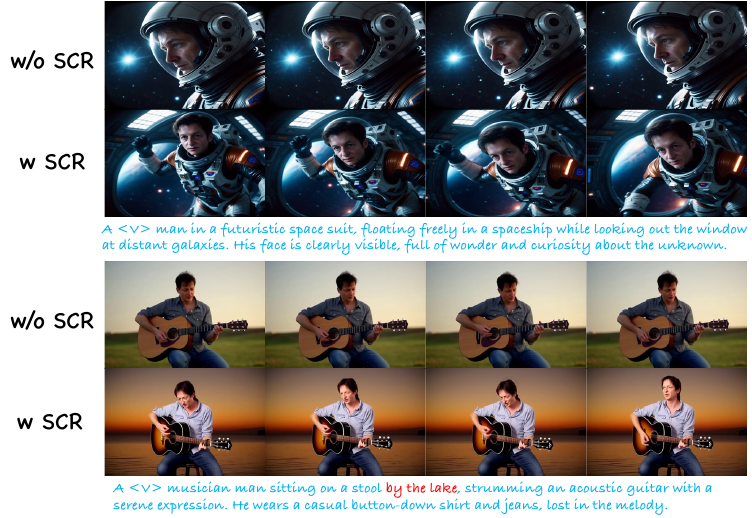}
\vspace{-0.3cm}
\caption{\textbf{Ablation study for the Semantic Consistency Reward.} As observed, tuning without SCR suffers from the dynamic degradation and the inferior semantic following, \ie, \textit{by the lake}.}
\label{fig:scr}
\vspace{-0.6cm}
\end{figure}

\subsection{Ablation Study}

\noindent
\textbf{Non-Reconstructive Training.} To validate the impact of proposed non-reconstructive training, we conduct a detailed ablation study in \cref{fig:text} and \cref{tab:icr}. They show the comparison between our PersonalVideo trained on the T2I model or T2V model, including with and without Prompt Augmentation. As observed, tuning on the T2I Model gets inferior ID fidelity due to the tuning-inference gap, which also exacerbates the distribution shift which leads to the blurred background. In contrast, tuning on the T2V model bridges the gap to achieve better ID fidelity with the preserved ability of semantic following.
On the other hand, tuning without simulated prompt augmentation overfits the reference images and disrupts the original capability of the semantic following, which manifests as an inability to precisely modify the background with reduced CLIP score. With the introduction of simulated prompt augmentation, this overfitting can be significantly reduced.

\noindent
\textbf{Semantic Consistency Reward.}
To demonstrate the effectiveness of our SCR loss, we conduct the experiments in \cref{fig:scr} and \cref{tab:scr}. As observed, the results of tuning without SCR loss suffer from the dynamic degradation in the first example, which exhibits less movements. Besides, the results has inferior semantic following in the second example, \ie, \textit{by the lake}. They verify that our proposed SCR effectively preserve the desired dynamics and semantics. 

\begin{figure}[t]
\centering
\includegraphics[width=.9\linewidth]{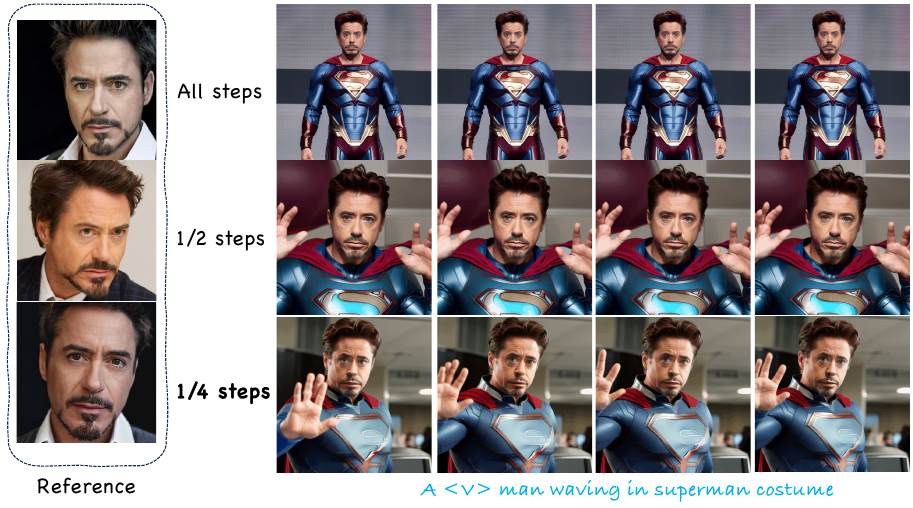}
\vspace{-0.3cm}
\caption{\textbf{Ablation for different steps to inject the identity.} As the denoising steps for injecting identity become more later, the motion dynamics of the generated videos improve accordingly.}
\label{fig:motion}
\vspace{-0.4cm}
\end{figure}

\noindent
\textbf{Different Steps to Inject the Identity.} We also conduct the ablation studies in \cref{fig:motion} and \cref{tab:motion}, which illustrate the improvement in motion dynamics of our \module to inject the identity only in the last quarter of denoising steps. As the denoising steps for injecting identity become more concentrated in the later stages, the motion dynamics of the generated videos improve accordingly. This aligns with our observations and validates the effectiveness of our design.

\vspace{-0.1cm}
\section{Conclusion \& Limitation}

In conclusion, we present \textbf{PersonalVideo}, a novel framework designed for identity-specific video generation, achieving high identity fidelity while preserving the motion dynamics and semantic following of the original T2V model. By applying direct reward supervision on generated videos, we successfully bridge the tuning-inference gap, mitigating the degradation. Furthermore, the simulated prompts augmentation enhances robustness, allowing for high-quality results even with minimal reference. Our method demonstrates superior performance over prior approaches, offering a flexible, efficient, and scalable solution for video personalization within the AIGC community.

However, our approach still has some limitations. While it enables a plug-and-play injection into the pre-trained T2V model, the results are inherently constrained by the capabilities of the T2V model itself. For example, it fails to generate customized videos that contain multiple identities. One possible solution is to further decouple the attention map of each subject, which will be explored in our future work.

{
    \small
    \bibliographystyle{ieeenat_fullname}
    \bibliography{main}
}

\clearpage
\maketitlesupplementary

\section{Implementation Details}
\label{sec:detail}
\textbf{Training Details.} For AnimateDiff, we use Stable Diffusion 1.5 with Realistic Vision~\citep{realisticvision2023} during training and inference. During training, we learn the \module for 800 iterations with a learning rate of 1e-4 with the batch size 1. We default to using AdamW optimizer with the default betas set to 0.9 and 0.999. The epsilon is set to the default 1e-8 and the weight decay is set to 1e-2.
For Hunyuanvideo, we employ the AdamW optimizer configured with a learning rate of 2e-5 and a weight decay parameter of 1e-4 for 4000 training steps.
During inference, we use 50 steps of DDIM sampler and classifier-free guidance with a scale of 7.5 for all baselines. We generate 16-frame videos with 512 × 512 spatial resolution for AnimateDiff and 61-frame videos with 720 × 1280 or 512 × 768 spatial resolution for HunyuanVideo. 
All experiments are conducted on a single NVIDIA A800 GPU. 

\noindent
\textbf{Baseline Details.} 
We compare our method with both optimization methods, such as MagicMe and Dreambooth-LoRA, and encoder-based methods such as IDAnimator and ConsisID. Specifically, Magic-Me is a recent T2V customization method that trains extended keywords on the Stable Diffusion and injects it into AnimateDiff. Besides, we compare with Dreambooth-LoRA, which uses traditional reconstructive loss during training. For a fair comparison, we train them for the same steps with PersonalVideo.
For the encoder-based methods, we compare with ID-Animator and ConsisID, the recent encoder-based methods, which are both trained on the large video dataset. 

\section{More Comparison}
\begin{figure*}[t]
\centering
\includegraphics[width=.9\linewidth]{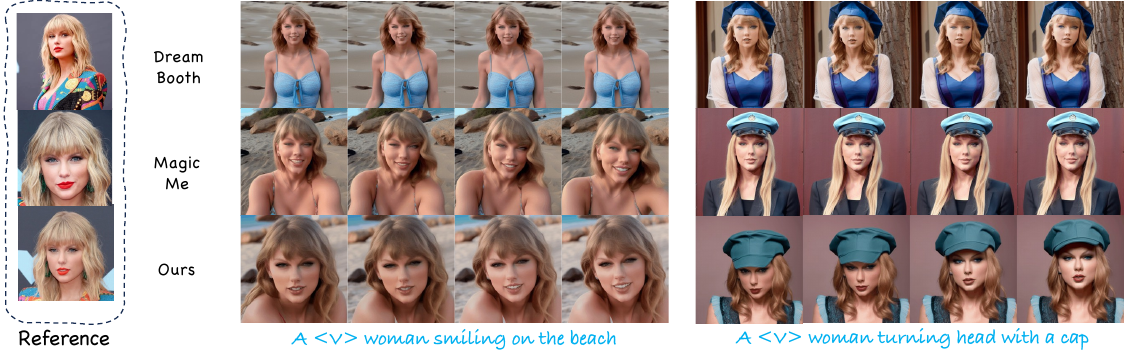}
\vspace{-0.3cm}
\caption{\textbf{Qualitative comparison on Animatediff.} As observed, both Dreambooth and MagicMe suffer from inferior ID fidelity. Besides, MagicMe has a semantic following degradation, \eg, \textit{tuning head}. In contrast, our PersonalVideo maintains high ID fidelity and preserve the original motion dynamics and semantic following, significantly surpassing others.}
\label{fig:result}
\vspace{-0.4cm}
\end{figure*}

\begin{figure*}[t]
\centering
\includegraphics[width=.9\linewidth]{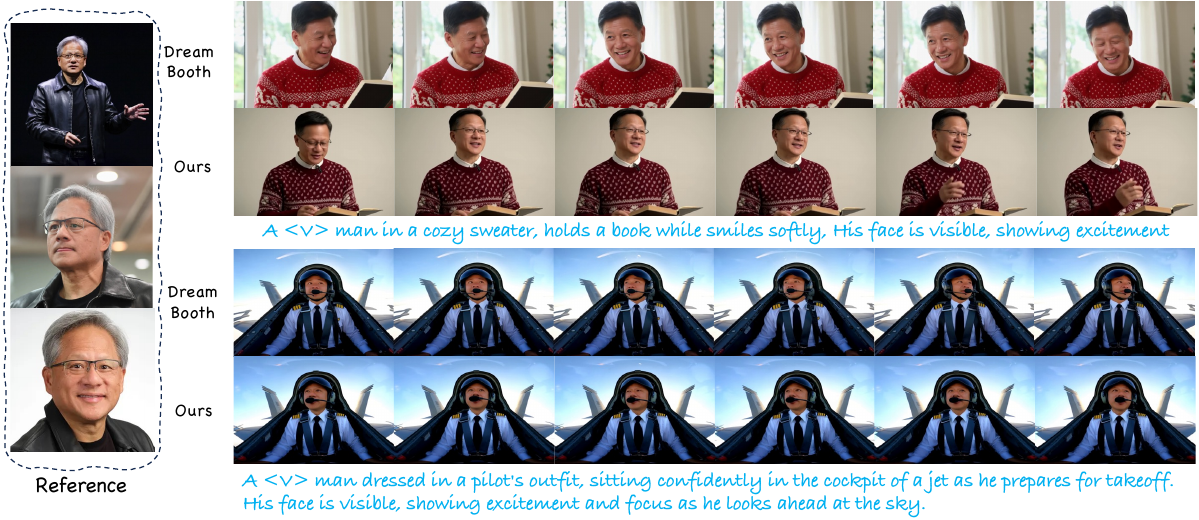}
\vspace{-0.3cm}
\caption{\textbf{More comparisons on HunyuanVideo.} As observed, Dreambooth suffers from inferior ID fidelity, while our PersonalVideo maintains higher ID fidelity without dynamic and semantic degradation, which is consistent with \cref{fig:result}.}
\label{fig:result2}
\vspace{-0.4cm}
\end{figure*}

\begin{figure*}[t]
\centering
\includegraphics[width=.9\linewidth]{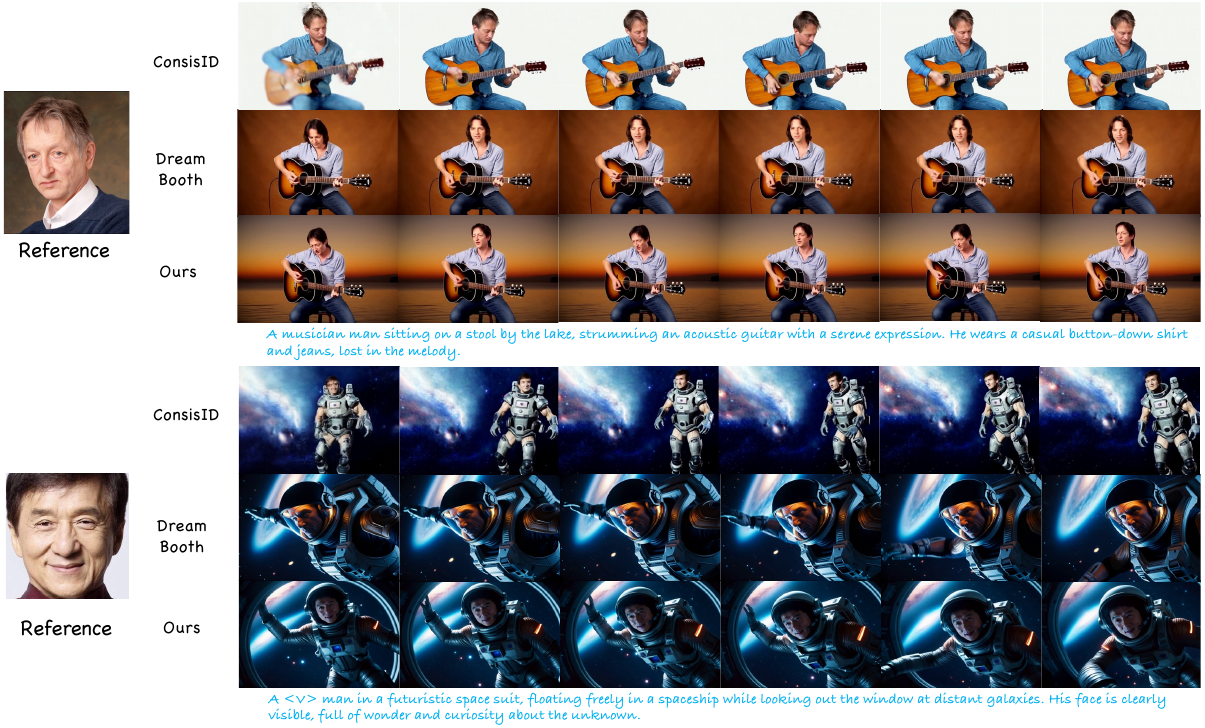}
\vspace{-0.1cm}
\caption{\textbf{More comparison for a single reference.} While ConsisID and Dreambooth suffer from the inferior ID fidelity, as well as severe degradation of motion dynamics and semantic following, \eg \textit{the stool by the lake}, our PersonalVideo achieves robust customization with high ID fidelity and preserved motion dynamics and semantic following.}
\label{fig:result3}
\vspace{-0.2cm}
\end{figure*}

We provide more comparison including more base models and different number of the references in \cref{fig:result}, \cref{fig:result2}, and \cref{fig:result3}.
As shown, both Dreambooth and MagicMe suffer from inferior ID fidelity. Besides, MagicMe has a severe misalignment of the prompt, \eg, \textit{tuning head}. In contrast, our PersonalVideo maintains higher ID fidelity without dynamic and semantic degradation, which is consistent with results on the DiT-based model.

\section{More ablation study}
\begin{figure*}[t]
\centering
\includegraphics[width=.9\linewidth]{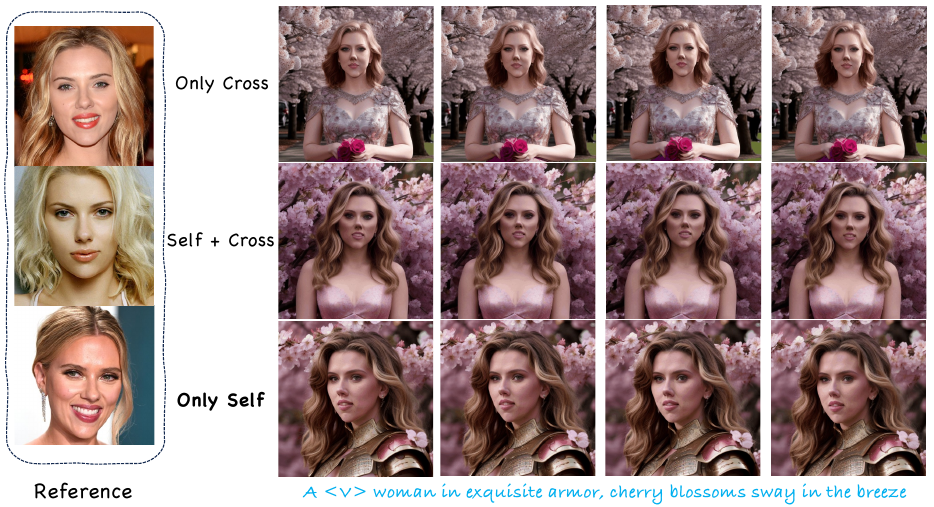}
\vspace{-0.3cm}
\caption{\textbf{Ablation for different layers to inject the identity.} As observed, injecting it on the cross-attention layer disrupts the ability of semantic following, \eg, the losing of \textit{exquisite armor}.}
\label{fig:self}
\vspace{-0.4cm}
\end{figure*}

\cref{fig:self} and \cref{tab:self} verify the improvement in semantic following of our Isolated Identity Adapter to inject the identity only on the spatial self-attention layer. As observed, injecting only on the cross-attention layer gets inferior ID fidelity with the reference images and disrupts the original capability of semantic following, such as the losing of \textit{exquisite armor}. Although injecting on both self-attention and cross-attention slightly achieves better ID fidelity, it still damages to the semantic following.

\begin{table}[t]
    \tablestyle{3pt}{1.15}
    \centering
    \begin{tabular}{lccc}
    \toprule
        & \textbf{Face} ($\uparrow$)  & \textbf{CLIP-T}  ($\uparrow$)& \textbf{Dynamic} ($\uparrow$)\\
            \midrule
        T2I w Aug &45.79 &28.42  &16.13\\
        T2V w/o Aug  &56.40 &24.10 &16.3\\
        \textbf{T2V w/ Aug} &\textbf{61.05} &\textbf{28.59} &\textbf{17.85}\\
    \bottomrule
    \end{tabular}
    \caption{Quantitative ablation of the non-reconstructive training.}
\label{tab:icr}
\vspace{-0.1cm}
\end{table}

\begin{table}[t]
    \tablestyle{3pt}{1.15}
    \centering
    \begin{tabular}{lccc}
    \toprule
        & \textbf{Face} ($\uparrow$)  & \textbf{CLIP-T}  ($\uparrow$)& \textbf{Dynamic} ($\uparrow$)\\
            \midrule
        w/o SCR &61.08 &26.38  &13.22\\
        \textbf{w/ SCR} &\textbf{61.05} &\textbf{28.59} &\textbf{17.85}\\
    \bottomrule
    \end{tabular}
    \caption{Quantitative ablation of Semantic Consistency Reward.}
\label{tab:scr}
\vspace{-0.1cm}
\end{table}

\begin{table*}[ht]
    \begin{subtable}{0.5\linewidth}
    \centering
    \tablestyle{3pt}{1.15}
    \begin{tabular}{cccc}
    \toprule
        & \textbf{Face} ($\uparrow$) & \textbf{Dynamic} ($\uparrow$) & \textbf{CLIP-T}($\uparrow$)\\
            \midrule
        All steps &62.37 &13.93  &26.95\\
        1/2 steps &60.36 &16.22 &25.63\\
        \textbf{1/4 steps} &\textbf{63.90} &\textbf{18.00} &\textbf{27.47}\\
    \bottomrule
    \end{tabular}
    \caption{Different steps to inject the identity.}
    \label{tab:motion}
     \end{subtable}
     \begin{subtable}{0.5\linewidth}
     \centering
     \tablestyle{3pt}{1.15}
    \begin{tabular}{cccc}
    \toprule
        & \textbf{Face} ($\uparrow$) &\textbf{CLIP-T} ($\uparrow$) & \textbf{Dynamic} ($\uparrow$) \\    \midrule
        Cross &42.68 &26.20 &17.70\\
        Self + Cross &\textbf{62.99} &23.35 &17.33\\
        \textbf{Self} &62.61 &\textbf{27.87} &\textbf{17.80}\\
    \bottomrule
    \end{tabular}
    \caption{Different layers to inject the identity in the Unet-based model.}
    \label{tab:self}
     \end{subtable}
    \vspace{-0.2cm}
     \caption{Quantitative ablation studies of Isolated Identity Adapter.
}
\vspace{-0.4cm}
\end{table*}

\section{More Results}
\begin{figure*}[t]
\centering
\includegraphics[width=\linewidth]{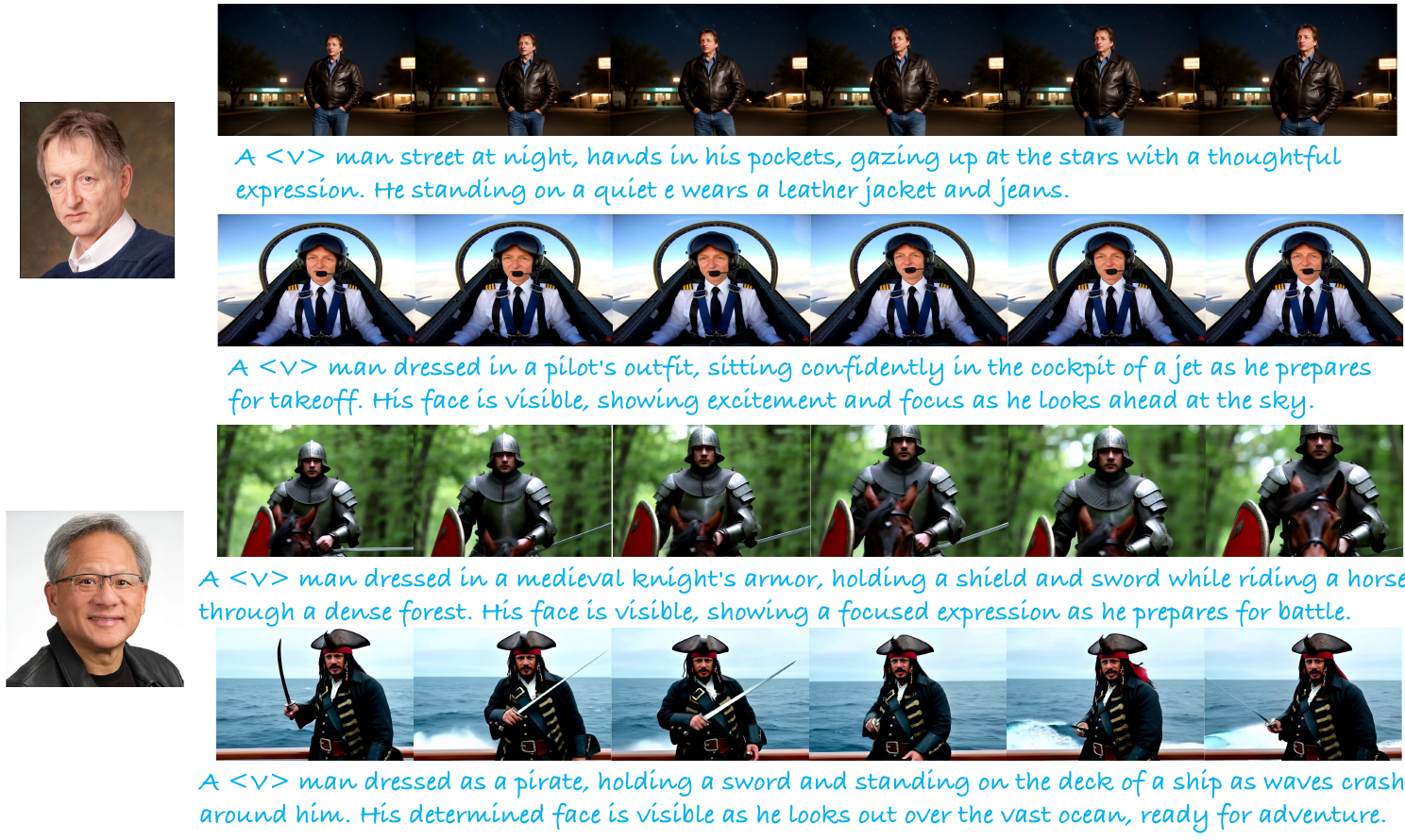}
\caption{\textbf{More results of PersonalVideo}.}
\label{fig:more1}
\end{figure*}

\begin{figure*}[t]
\centering
\includegraphics[width=\linewidth]{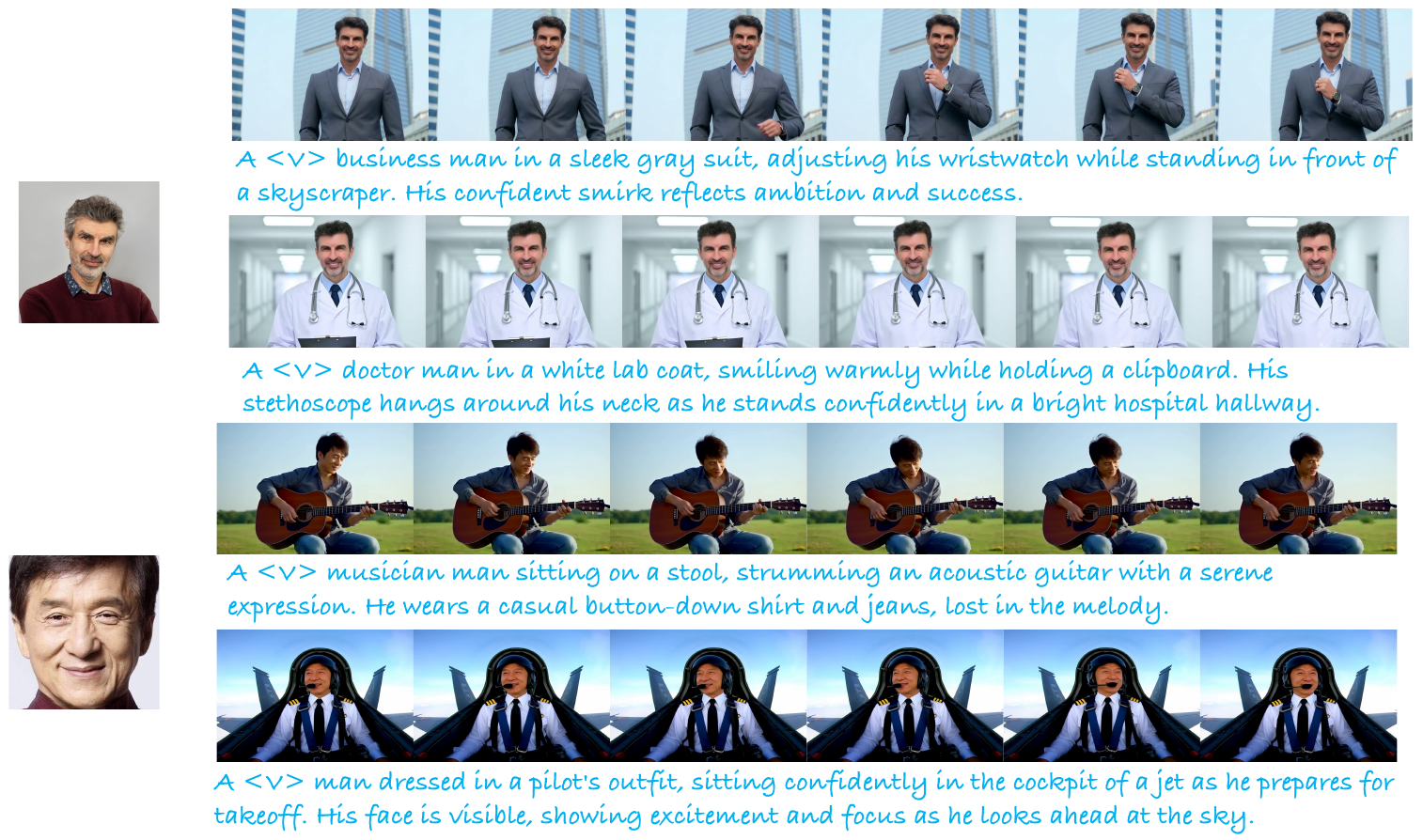}
\caption{\textbf{More results of PersonalVideo}.}
\label{fig:more2}
\end{figure*}

\begin{figure*}[t]
\centering
\includegraphics[width=\linewidth]{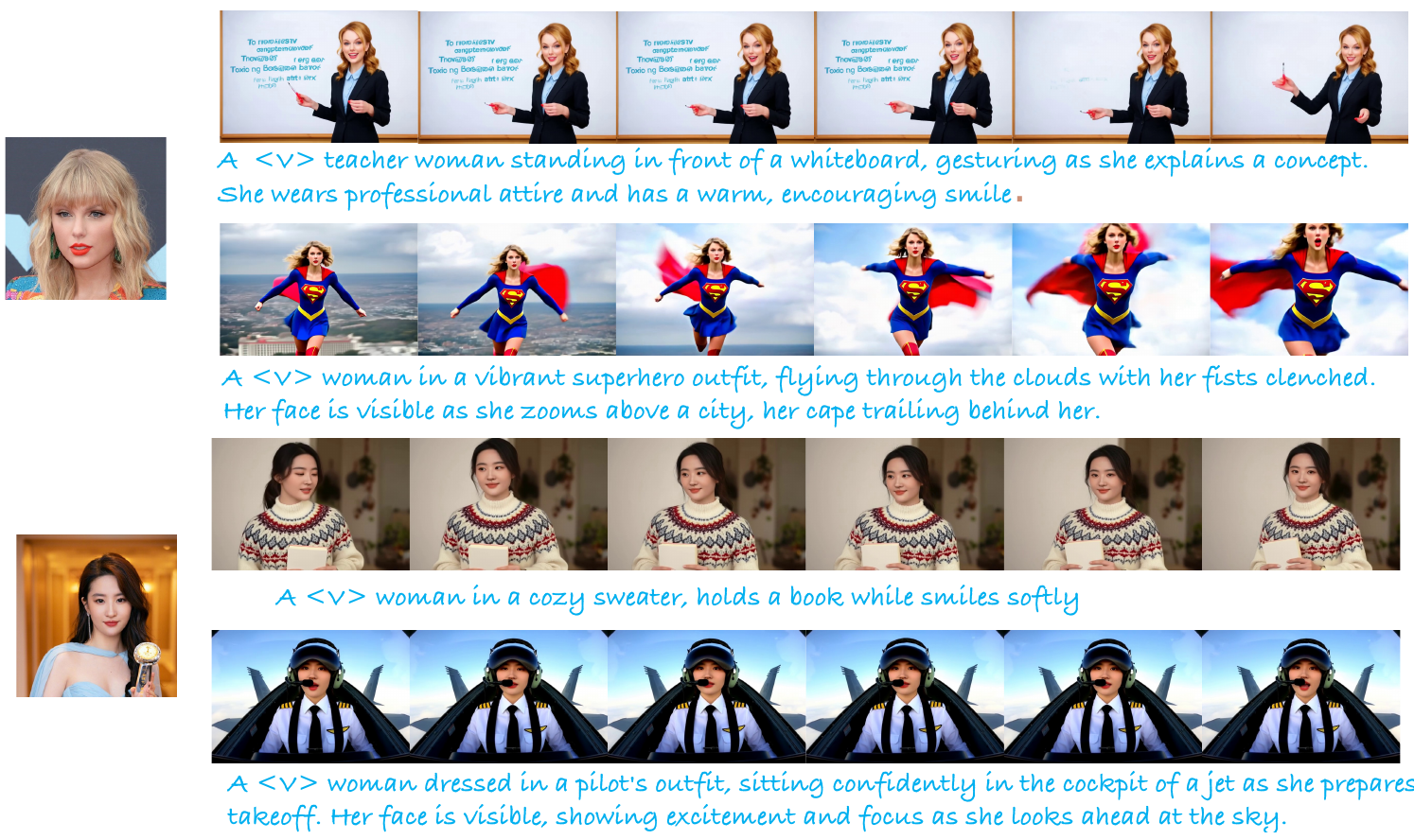}
\caption{\textbf{More results of PersonalVideo}.}
\label{fig:more3}
\end{figure*}

\begin{figure*}[t]
\centering
\includegraphics[width=\linewidth]{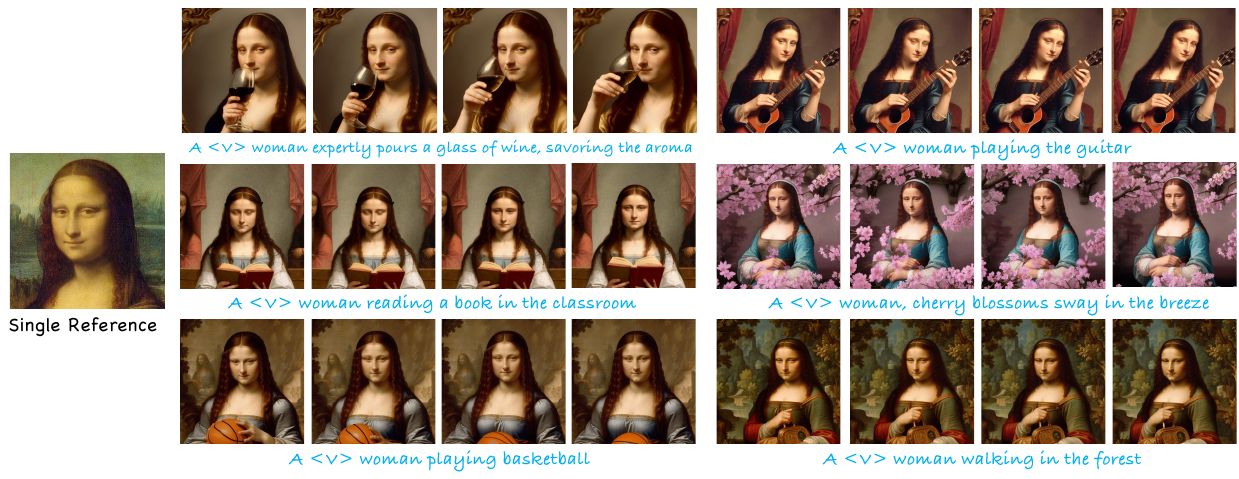}
\caption{\textbf{More results of PersonalVideo} with only just one image.}
\label{fig:more_single}
\end{figure*}

\begin{figure*}[t]
\centering
\includegraphics[width=.9\linewidth]{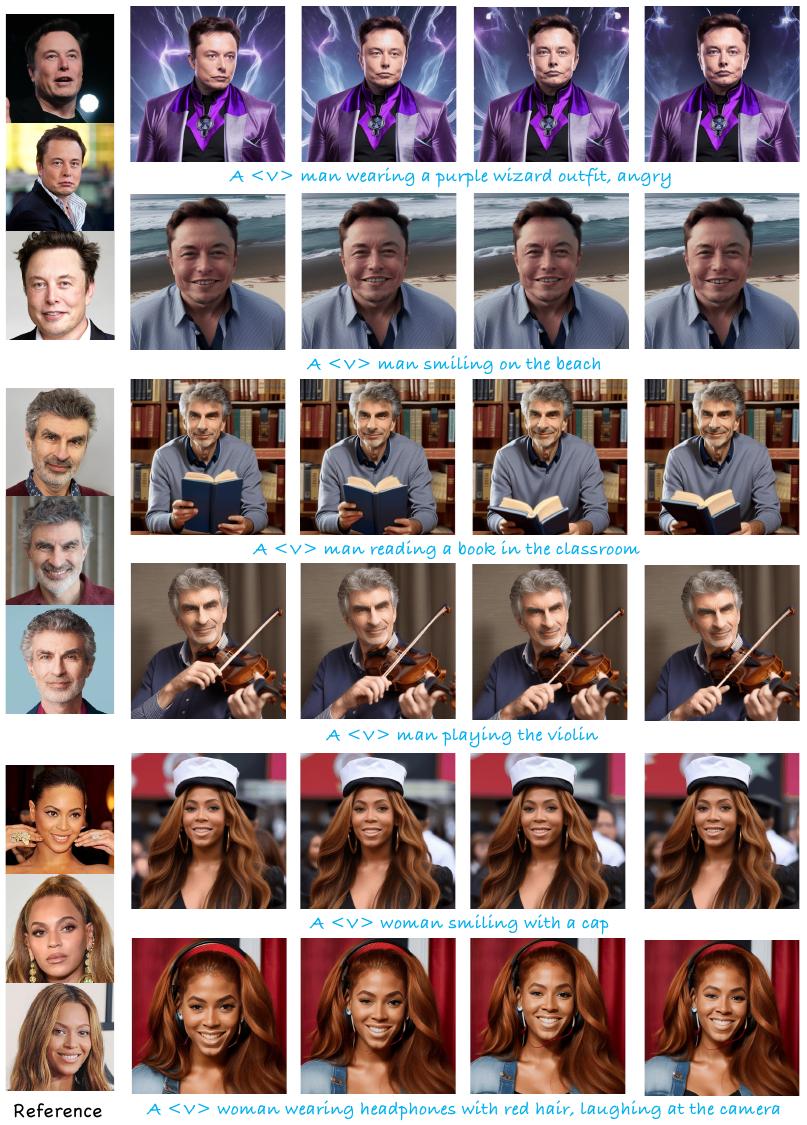}
\caption{\textbf{More results of PersonalVideo} with few images.}
\label{fig:more}
\end{figure*}

As shown in \cref{fig:more1}, \cref{fig:more2}, \cref{fig:more3}, \cref{fig:more_single}, and \cref{fig:more}, we present more customization results of PersonalVideo, including few or just one reference image. They showcase it achieves high ID fidelity and preserves original motion dynamics and semantic following, which provides further evidence of its promising performance and robustness.

\section{Reproducibility Statement}
We make the following efforts to ensure the reproducibility of PersonalVideo: (1) Our training and inference codes together with the trained model weights will be publicly available. (2) We provide training details in the appendix (\cref{sec:detail}), which is easy to follow. (3) We provide the details of the human evaluation setups in the appendix (\cref{sec:user}).

\section{Impact Statement}
Our main objective in this work is to empower novice users to generate visual content creatively and flexibly. However, we acknowledge the potential for misuse in creating fake or harmful content with our method. Thus, we believe it's essential to develop and implement tools to detect biases and malicious use cases to promote safe and equitable usage.

\begin{figure*}[t]
\centering
\includegraphics[width=\linewidth]{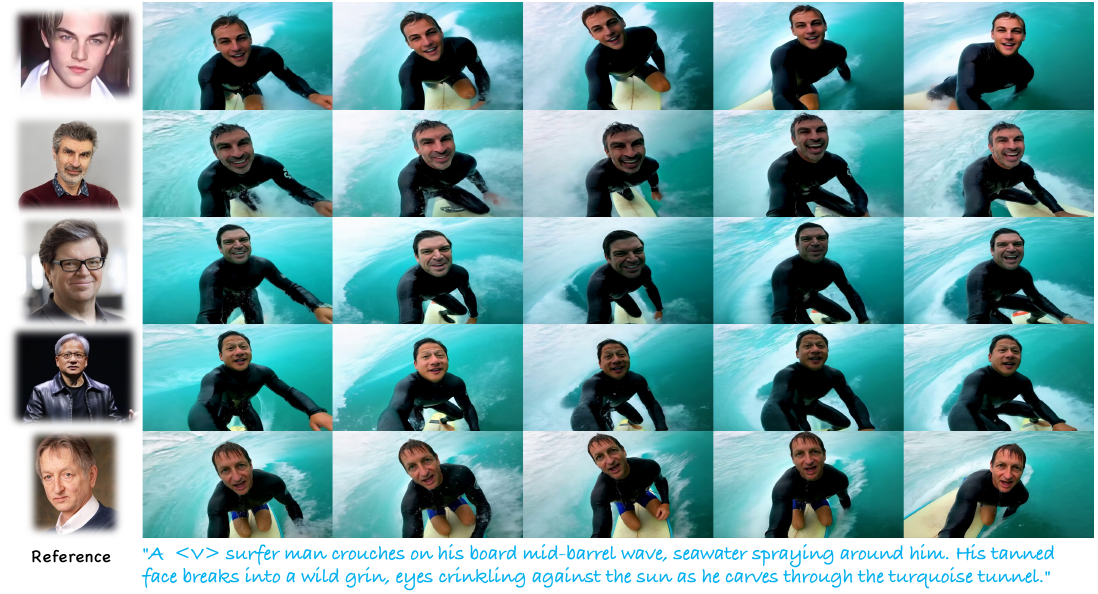}
\caption{\textbf{More results of PersonalVideo}.}
\label{fig:a2}
\end{figure*}

\begin{figure*}[t]
\centering
\includegraphics[width=\linewidth]{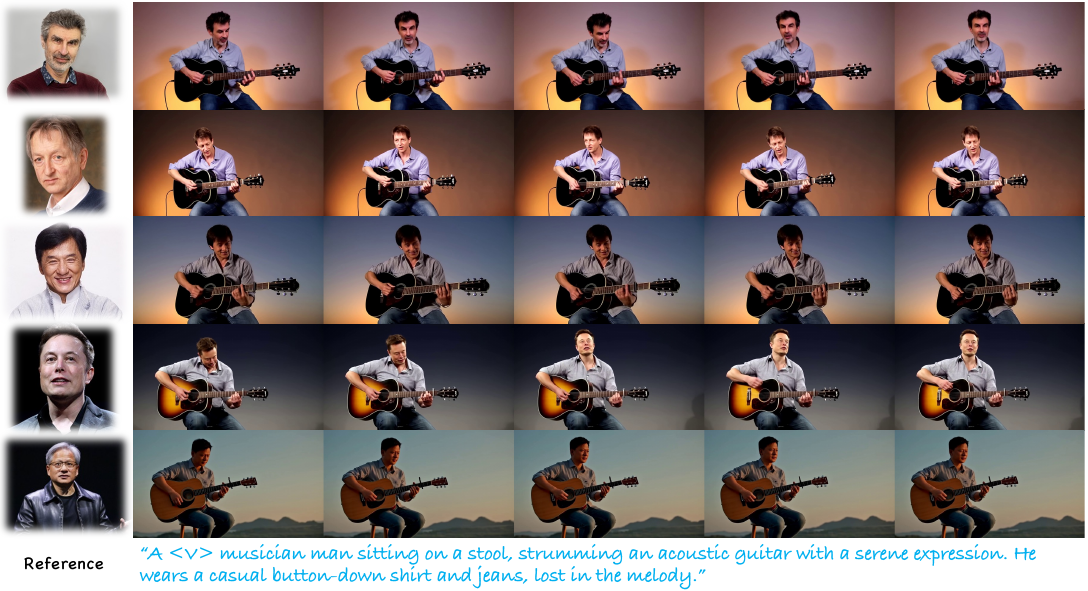}
\caption{\textbf{More results of PersonalVideo}.}
\label{fig:a3}
\end{figure*}

\begin{figure*}[t]
\centering
\includegraphics[width=\linewidth]{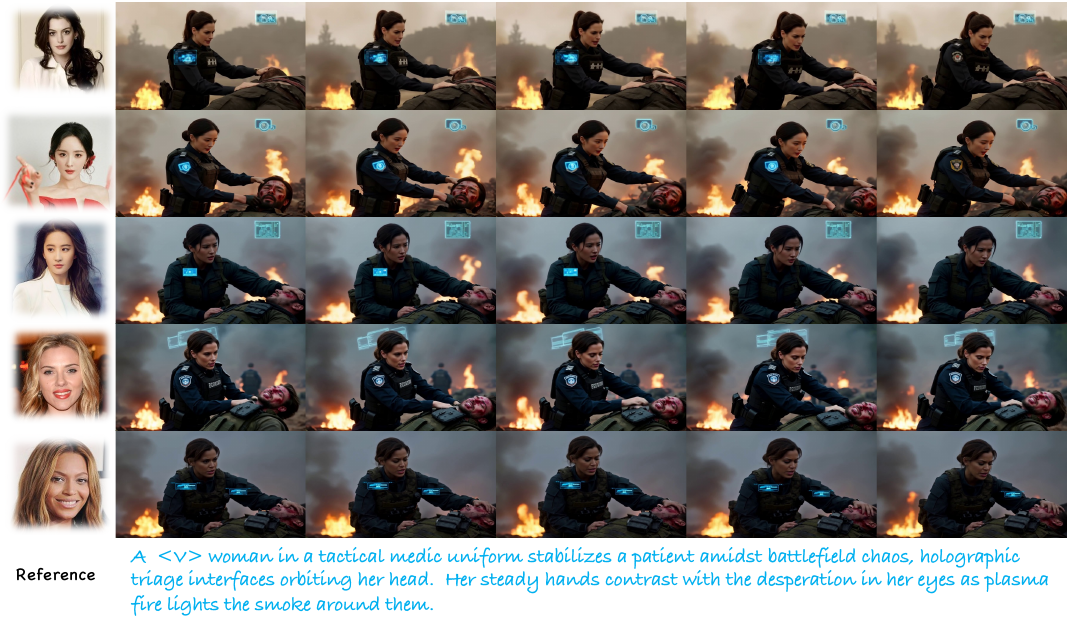}
\caption{\textbf{More results of PersonalVideo}.}
\label{fig:a4}
\end{figure*}

\begin{figure*}[t]
\centering
\includegraphics[width=\linewidth]{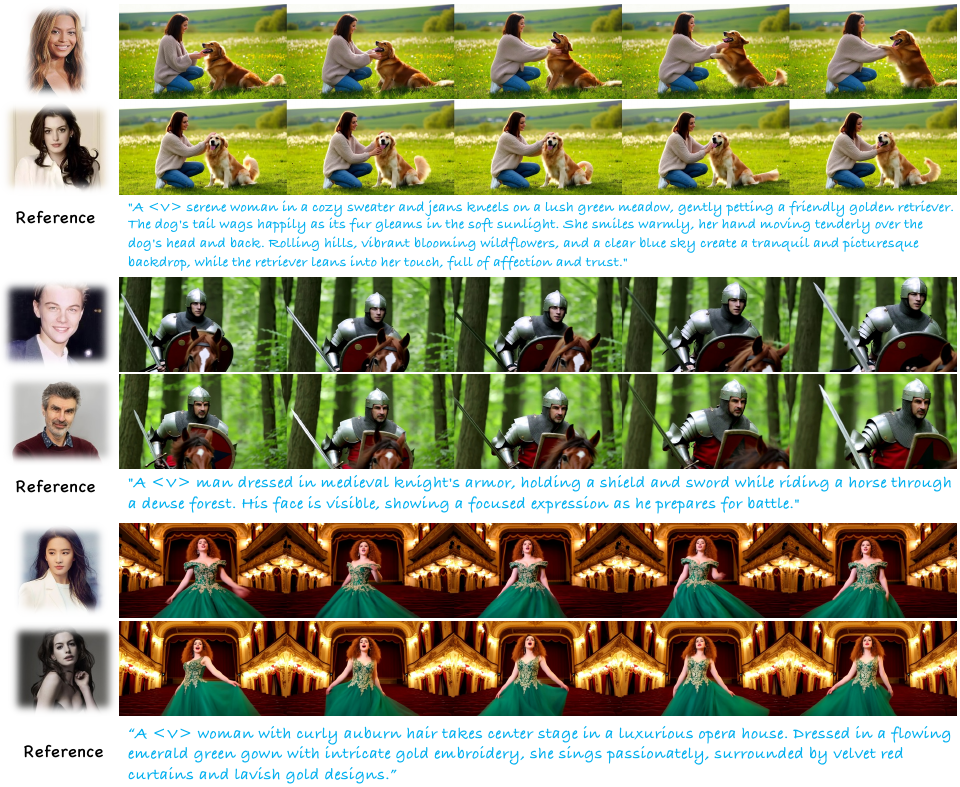}
\caption{\textbf{More results of PersonalVideo}.}
\label{fig:a1}
\end{figure*}

\end{document}